\documentclass{article}
\usepackage{spconf,amsmath,graphicx}

\title{Automated segmentaiton and classification of arterioles and venules using Cascading Dilated Convolutional Neural Networks}
\name{Meng Li$^{1}$, Yan Zhang$^{3}$, Haicheng She$^{4, 5}$,  Jinqiong Zhou$^{4, 5}$, Jia Jia$^{3}$, Danmei He$^{3}$, Li Zhang$^{\star, 1, 2}$\thanks{$^{\star}$Corresponding author: Li Zhang, zhangli\underline{ }pku@pku.edu.cn.}}
\address{$^{1}$ Center for Data Science, Peking University, Beijing, China; 
	\\$^{2}$ Beijing Institute of Big Data Research, Beijing, China; 
	\\$^{3}$ Department of Cardiology, Peking University First Hospital, Beijing, China;
	\\$^{4}$ Beijing Tongren Eye Center, Beijing Tongren Hospital, Capital Medical University, Beijing, China;
    \\$^{5}$ Beijing Ophthalmology and Visual Science Key Laboratory, Beijing, China.}
\begin{document}
%
\maketitle
\begin{abstract}
The change of retinal vasculature is an early sign of many vascular and systematic diseases, such as diabetes and hypertension. Different behaviors of retinal arterioles and venules form an important metric to measure the disease severity. Therefore, an accurate classification of arterioles and venules is of great necessity. In this work, we propose a novel architecture of deep convolutional neural network for segmenting and classifying arterioles and venules on retinal fundus images. This network takes the original color fundus image as inputs and multi-class labels as outputs. We adopt the encoding-decoding structure (Unet) as the backbone network of our proposed model. To improve the classification accuracy, we develop a special encoding path that couples InceptionV4 modules and Cascading Dilated Convolutions (CDCs) on top of the backbone network. The model is thus able to extract and fuse high-level semantic features from multi-scale receptive fields. The proposed method has outperformed the previous state-of-the-art method on DRIVE dataset with an accuracy of 0.955 $\pm$ 0.002.
\end{abstract}
\section{Introduction}
\label{sec:intro}

The changes of retinal vasculature contain substantial diagnostic information for many vascular and systematic diseases. Specifically, diseases may affect arterioles and venules differently. For example, in hypertensive patients, the size of arterioles usually shrinks faster than that of venules. While in diabetic patients, we usually observe the expansion of venules first. Therefore, accurately segmenting and classifying retinal arterioles and venules has a great potential to improve the diagnosis and management of these diseases.

Methods for segmenting and classifying retinal arterioles and venules include two major types: image-processing methods \cite{xu2017improved,kondermann2007blood} and deep learning based methods \cite{welikala2017automated,albadawi2018arterioles}. For image-processing methods, a vessel segmentation-arteriovenous classification strategy is usually adopted. A binary mask is first generated by retinal vessel segmentation of the fundus image. Furthermore, the centerlines of the vessels are computed from the binary mask using image morphological operations. Various hand-crafted image features are then extracted around vessel centerlines, which are used to classify the arterioles and the venules in the image. 

On the other hand, deep learning based methods are reported to classify arterioles and venules. Welikala et al \cite{welikala2017automated} present a two-stage method for automated classification of arteriole and venule using deep learning. Retinal vessels are first segmented using a linear detector. The centerlines of arterioles and venules are then classified by a 6-layer neural network. AlBadawi et al. \cite{albadawi2018arterioles} report a deep learning method that combines an encoding-decoding model and graph-based approach to classify arterioles and venules. These methods substantially improve segmentation accuracies as opposed to traditional image-understanding approaches, but the pipelined workflow may undermine the stability of the methods. 

We propose a novel architecture of deep convolutional neural network for segmenting and classifying arterioles and venules on retinal fundus images. At beginning, we first adopt the encoding-decoding structure (Unet) as the backbone network of our proposed model. However, the model generates poor segmentation and classification results with the classic convolutional layers in Unet. One explanation of this problem is that retinal vessels in fundus images strictly follow a topological distribution: the same type of blood vessels does not intersect itself. The fixed receptive fields of classic convolutional layers are insufficient to represent such global image information. Therefore, to improve the accuracies of segmentation and classification, we develop a special encoding path that couples InceptionV4 modules and Cascading Dilated Convolutions (CDCs) on top of the backbone network. The model is thus able to extract and fuse high-level semantic features from multi-scale receptive fields. The network structure is shown in Fig. 1.

This network takes the original color fundus image as inputs and multi-class labels as outputs, which follows an end-to-end training process, requiring limited pre- and post-processing of the image data. All the image features are computed and utilized internally in the deep neural network, where no hand-crafted features or task-specific assumptions are included. The proposed method is evaluated on the DRIVE dataset \cite{staal2004ridge} and has achieved state-of-the-art performance.

\begin{figure}[!htp]
	\centering
	\centerline{\includegraphics[width=0.95\linewidth]{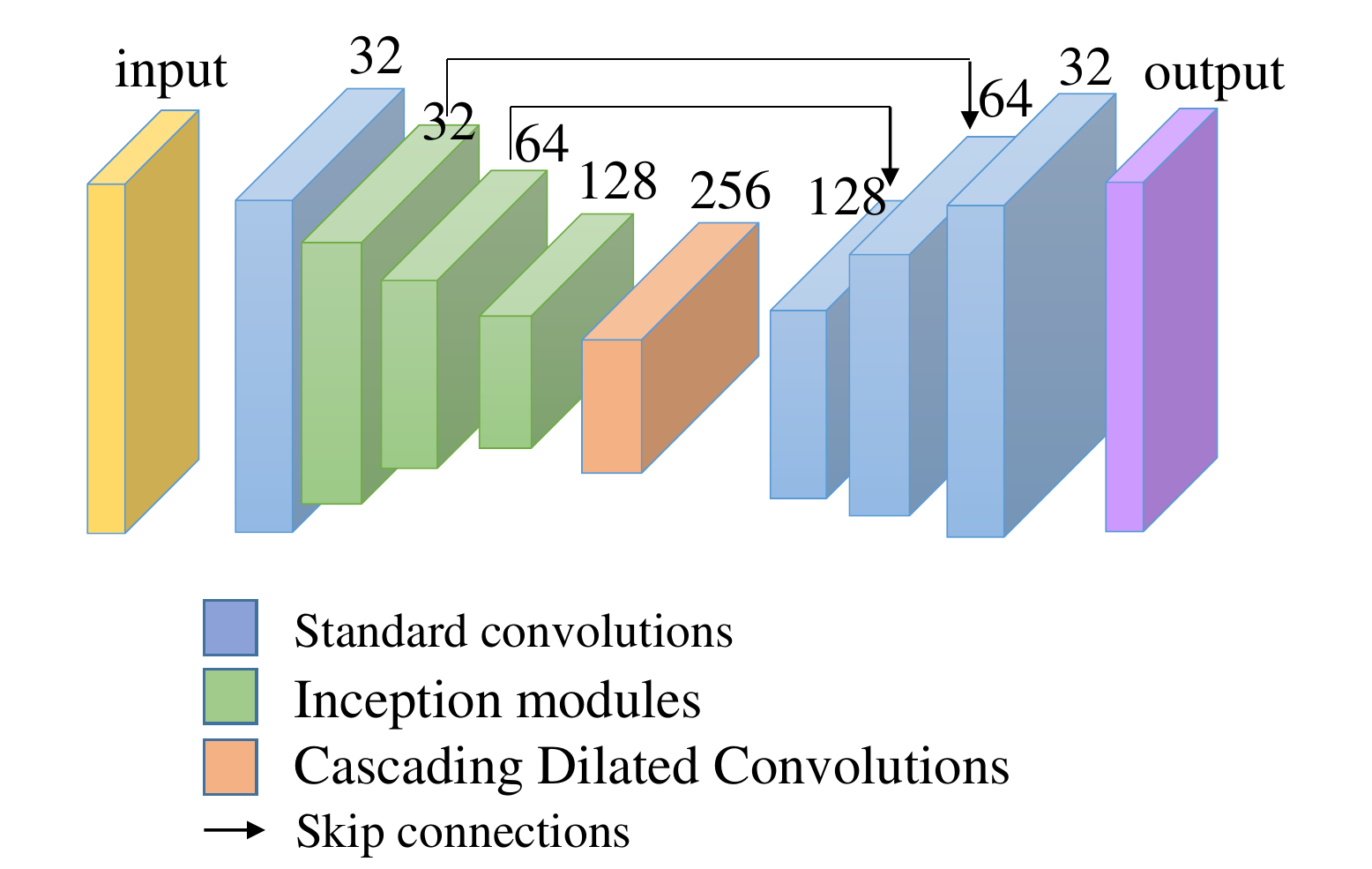}}
	\caption{The overall architecture of the proposed model.}
	\label{fig:overall}
\end{figure}

\section{Methods}
\label{sec:methods}

We take the encoding-decoding structure (Unet \cite{ronneberger2015u}) as the backbone network of our model. In the encoding stage, we use Inception convolutional modules to extract and fuse high-level features via multi-layer feature extraction. To further enhance the ability of information collection, we implement Cascading Dilated Convolutions (CDCs) to extract features with enlarged receptive fields. In the decoding stage, a step-by-step upsampling process restores image resolution. Skip connections between the encoding and decoding paths allow the information directly flows through corresponding layers, which maintains the information magnitude, avoiding gradient vanishing during training process. Fig. \ref{fig:overall} shows the overall architecture of our proposed model.

\begin{figure}[!ht]
	\centering
	\centerline{\includegraphics[width=0.95\linewidth]{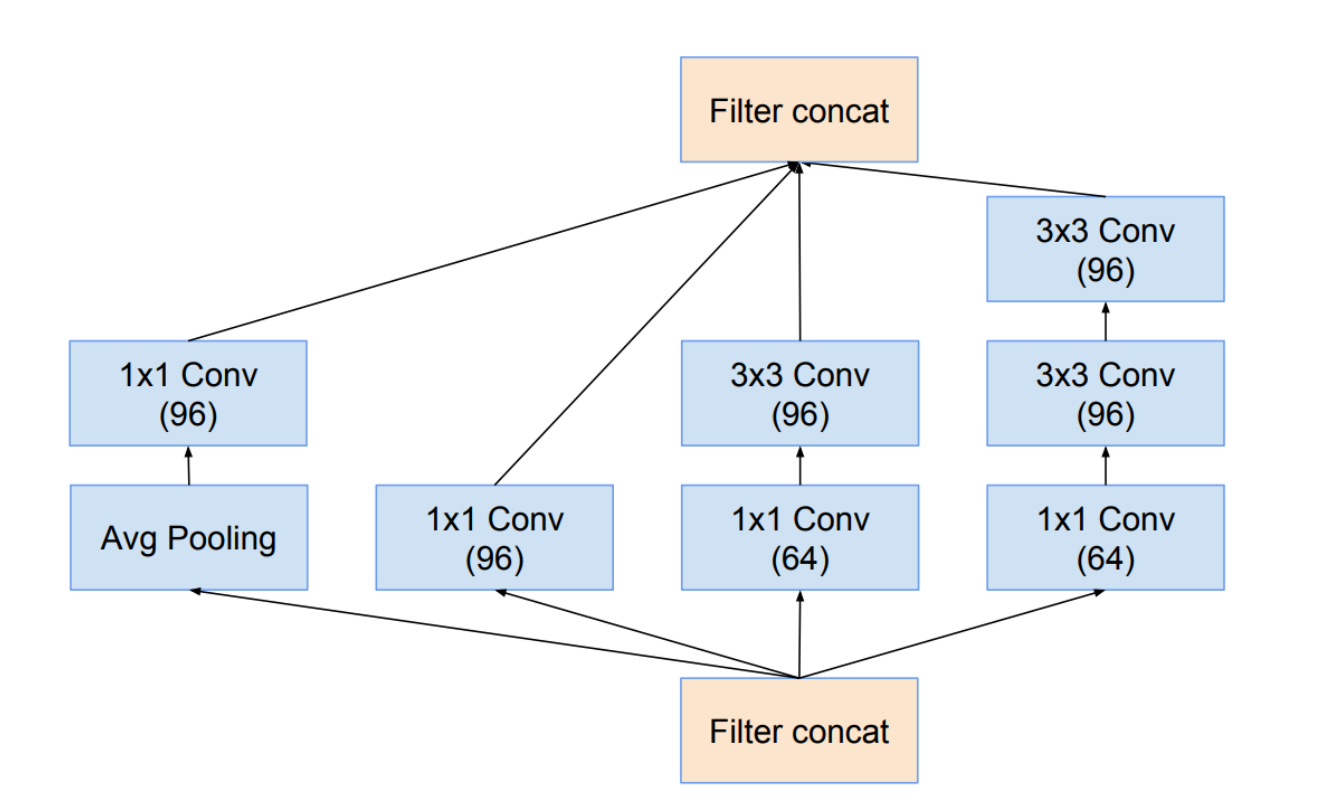}}
	\caption{An illustration of Inception convolutional module.}
	\label{fig:inception}
\end{figure}

\subsection{Inception convolutional module}
In the fundus images, retinal vasculature is complex due to the irregular distribution of vessel bifurcations and intersections. We thus adopt Inception convolution module in \cite{szegedy2017inception} for the feature extraction of retinal arterioles and venules. As shown in Fig. \ref{fig:inception}. Inception module comprises of a convolutional block and a down-sampling block. The convolutional block contains multiple branches, where each branch first uses the 1x1 convolution kernel to reduce the number of feature channels, and the convolutions with kernels of 3x3, 1x7, and 7x1 are then used to compute feature maps from different scales. Finally, multi-scale information is grouped by concatenating all feature maps. The down-sampling block is a concatenation of max-pooling layer and convolution kernel with stride of 2 to resample the input/intermediate feature maps. Compared the traditional max-pooling, this new structure could effectively avoid information loss. 

\begin{figure}[!ht]
	\centering
	\centerline{\includegraphics[width=0.95\linewidth]{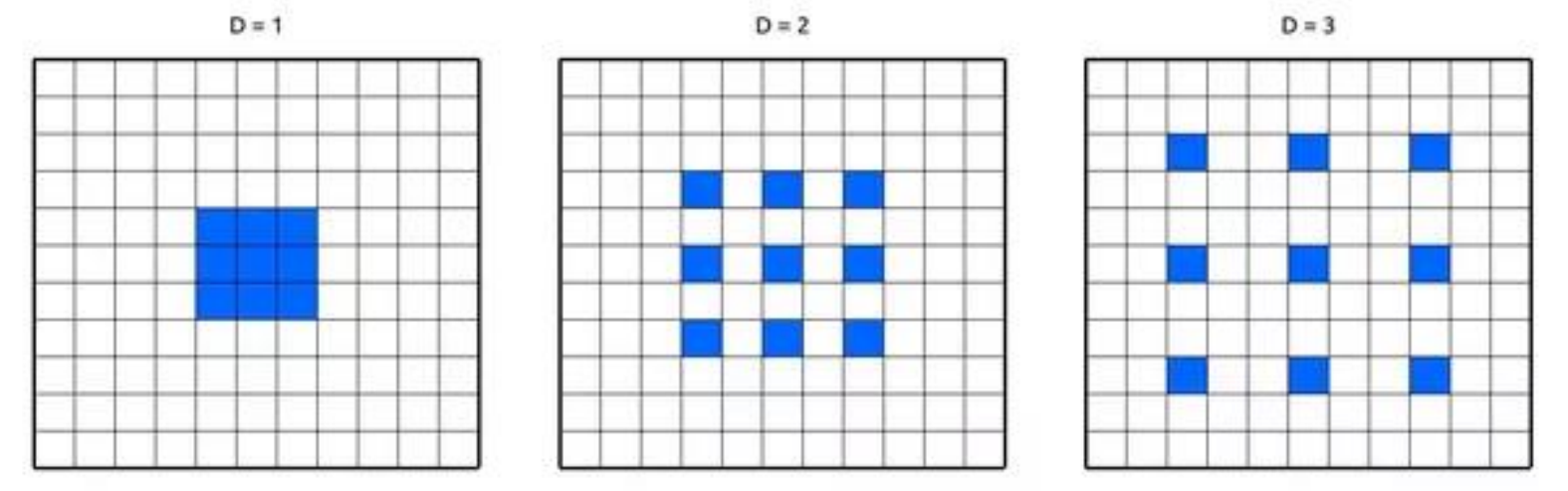}}
	\caption{Dilated convolution: insert zeros into classic convolution to increase the size of receptive fields.}
	\label{fig:dilated}
\end{figure}

\subsection{Cascading Dilated Convolutions}
In the arteriovenous classification of retinal fundus images, vessels usually have long and curvy shapes. However, the small receptive field in classic convolution is designed for relatively local features and works poorly for classifying most of the main branches of retinal arterioles and venules. Therefore, we adopt dilated convolution which is described in \cite{chen2018deeplab} (see Fig. \ref{fig:dilated}). The authors combines multiple convolutions with different dilation rates in a parallel fashion, extracting features from different scales by using enlarged receptive fields. Different from their original parallel design for segmenting multiple objects with different sizes, we implement a novel structure with cascading dilated convolutions, called CDCs. The cascading design suits the continuous structures of retinal vessel trees which have thick trunks and thin branches. The proposed CDCs combines four dilated convolution layers of rates 2, 4, 8, and 12, respectively, which represent different sizes of receptive fields. CDCs is shown in Fig.6.

\begin{figure}[!ht]
	\centering
	\centerline{\includegraphics[width=0.95\linewidth]{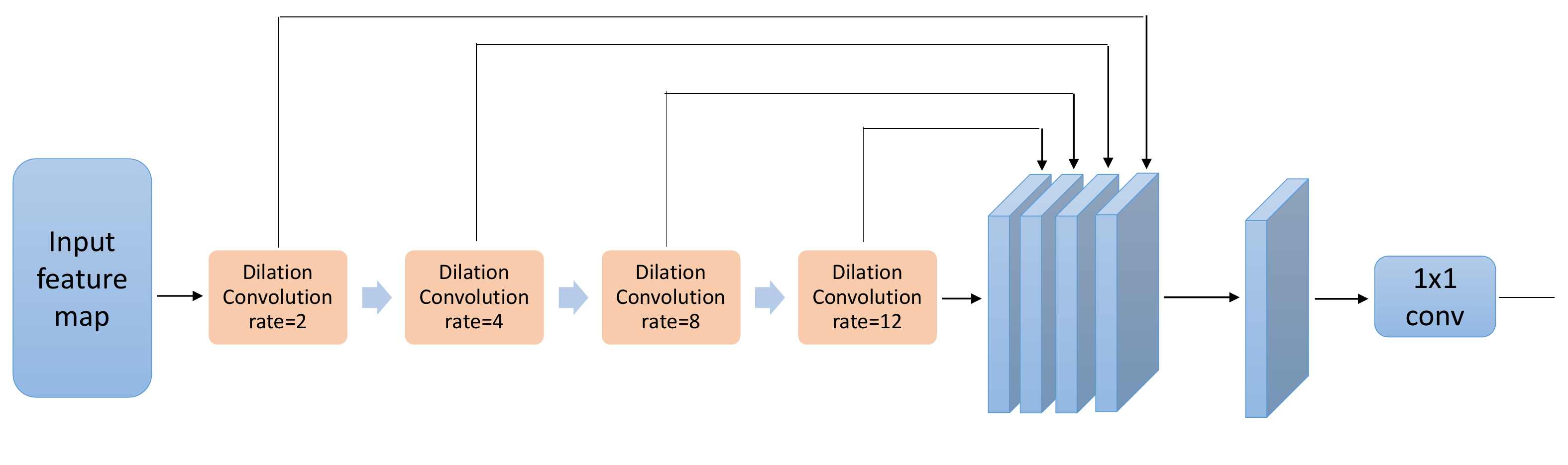}}
	\caption{An illustration of Cascading Dilation Convolutions.}
	\label{fig:cdcs}
\end{figure}

\section{Experiments and Results}
\label{sec:experiments}

\subsection{Data}
We conduct our experiment on a publicaly available dataset, the DRIVE dataset. DRIVE dataset contains 40 retinal fundus images with the annotations of arterioles and venules. An example of retinal fundus images and annotations is shown in Figure XXX. The image size is 584*565. We use a randomly selected subset of 30 images for training the deep neural network and the rest 10 images to evaluate the results. Due to the small size of this dataset, the images are first randomly cropped to a size of 512*512, and scaling, panning, random clipping, and etc are then applied to augment the dataset. Finally, the size of training dataset is expended to 2,490 images. A case-level five-fold cross validation is performed within the training dataset. In order to balance the data imbalance between arteriovenous and background pixels, we set the class weight to 5 for veins and arteries during the experiment.

\begin{figure}[!ht]
	\centering
	\centerline{\includegraphics[width=0.95\linewidth]{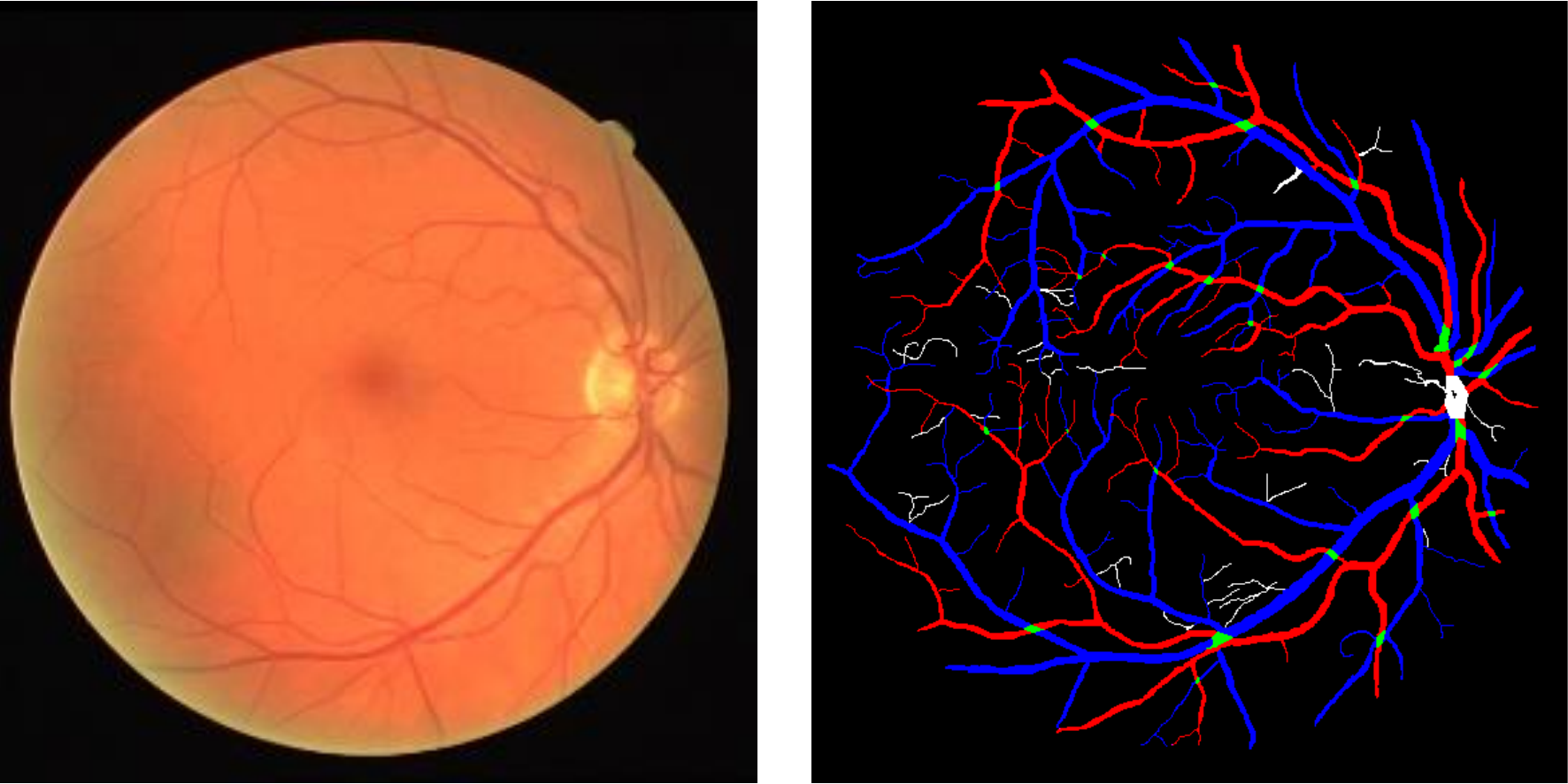}}
	\caption{Left: original image; Right: ananotated image.}
	\label{fig:two}
\end{figure}

As shown in Fig. \ref{fig:two}, five labels are provided by the anotations which are red for arterioles, blue for venules, green for intersections, black for background and white for uncertain pixels. In the experiment, we remove the label of uncertain pixels, resulting in four classes/labels for the model training and predicting. Furthermore, the intersections have textural characteristics from both arterioles and venules, we thus set a very small class weight (1e-12) to the intersection to avoid ambiguity during training. 

\subsection{Model Regularization}
To prevent the possible overfitting issue from training a small dataset, we apply the following approachs:
(1) Add Batch Normalization layer after each layer of convolution. (2) In the encoding stage, a dropout operation is added after each convolution. The threshold of dropout is set to 0.2, respectively. (3) Reduce the number of parameters in the convolution layer. In encoding stage, the numbers of channels are set to 32, 32, 64, 128, respectively. The convolution number in Decoder stage is set to 128, 64, and 32, respectively.

\subsection{Experimental results}
The model is trained by standard backpropagation and gradient descent approaches. The learning rate is initialized at 1e-4, and to increase the speed of convergence, poly decay \cite{liu1506parsenet} is used as the learning rate decay strategy. We use cross-entropy between the prediction and the label as loss function. During the experiment, the batch size is set to 4, and two NVIDA 1080 GPUs are used for parallel training.

\begin{figure}[!htp]
	\centering
	\centerline{\includegraphics[width=0.95\linewidth]{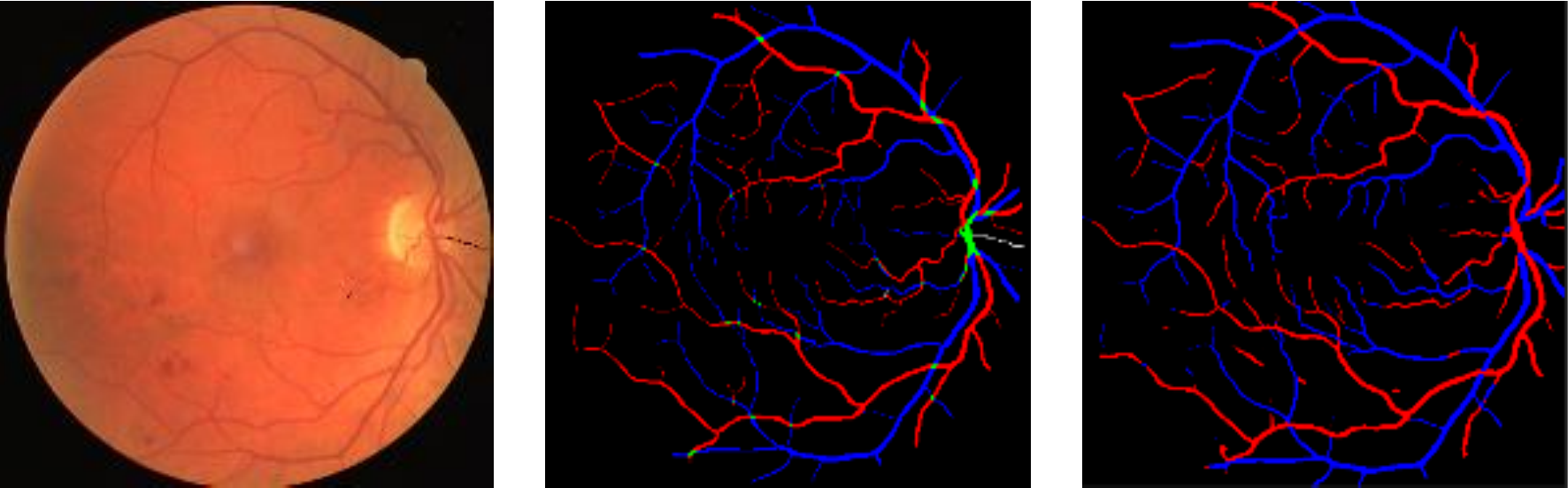}}
	\caption{Left: original image, Middle: annotated image, Right: classification result. Red represents arterioles and blue represents venules. The uncertain pixels (white) are not classified, and the intersections (green) are automatically classified as arterioles or venules according due to a small class weight.}
	\label{fig:three}
\end{figure}

We use true positive rate (TPR) and accuracy to evaluate the classification results of the proposed model.
\begin{equation}
TPR_{at} = \frac{TP_{at}}{TP_{at}+FP_{at}}
\end{equation}
\begin{equation}
TPR_{ve} = \frac{TP_{ve}}{TP_{ve}+FP_{ve}}
\end{equation}
\begin{equation}
Accuracy=\frac{TP_{ve}+TP_{at})}{(TP_{ve}+FP_{ve}+TP_{at}+FP_{at})}
\end{equation}
where the subscripts at and ve refer to arterioles and venules, respectively. TP stands for true positive and FP stands for false positive, as described in \cite{xu2018simultaneous}. The experimental results are shown in Table \ref{tab:table2}. Fig. \ref{fig:three} shows an example of the segmentation and classification results.

\begin{table}[!t]
		\begin{tabular*}{\linewidth}{c c}
			\hline
			Methods & Accuracy  \\ 
			\hline
			Xu et al. \cite{xu2016smartphone} &0.933 \\
			Zhang et al. \cite{zhang2016robust} &0.948  \\
			Li et al. \cite{li2016cross} &0.953 \\
			\textbf{Proposed Method} &\textbf{0.955} \\
			\hline
	\end{tabular*}
	\caption{Comparison between the proposed method and previous methods.}
	\label{tab:table2}
\end{table}

\section{CONCLUSION}
\label{sec:conclusion}

In this work, we present a novel deep-learning framework for automated segmentation and classification of arterioles and venules. We adopt Unet (encoding-decoding) structure as our backbone network, which provides relatively good baseline of this work. Furthermore, we exploit the Inception convolution module to extract and fuse high-level features. During the experiment, we discover that the classic convolution suffers ineffective information collection for complex vascular structures, due to the small receptive fields. To address this, we develop a novel structure named cascading dilated convolution (CDC), which progressively enlarge the size of receptive fields to segment and classify arterioles and venules. Finally, several methods are applied to prevent the possible overfitting during the model training. The experimental results demonstrate the superior performance of the proposed method.

\bibliographystyle{IEEEbib}
\bibliography{refs}

\end{document}